\documentclass{article}
\usepackage{ijcai25}

\usepackage{times}
\usepackage{soul}
\usepackage{url}
\usepackage[hidelinks]{hyperref}

\hypersetup{
    colorlinks=true,
    linkcolor=red, 
    citecolor=blue, 
    urlcolor=magenta    
}

\usepackage[utf8]{inputenc}
\usepackage[small]{caption}
\usepackage{graphicx}
\usepackage{amsmath}
\usepackage{amsthm}
\usepackage{booktabs}
\usepackage{algorithmic}
\usepackage[ruled, vlined]{algorithm2e}
\usepackage[switch]{lineno}

\usepackage{amssymb, amsmath}
\usepackage{bm}
\usepackage{xcolor}

\title{RLMiniStyler: 
Light-weight RL Style Agent for \\ Arbitrary Sequential Neural Style Generation}

\author{
Jing Hu$^1$
\and
Chengming Feng$^1$
\and
Shu Hu$^2$
\and
Ming-Ching Chang$^3$
\and
Xin Li$^3$
\and
Xi Wu$^{1}$,
Xin Wang$^{3}$\thanks{Corresponding Author} 
 \\
\affiliations
$^1$Chengdu University of Information Technology \\ 
$^2$Purdue University \\ 
$^3$University at Albany, SUNY \\
\emails
{\tt {\small jing\_hu09@163.com,
fengxiaoming520@gmail.com,
hu968@purdue.edu,
mchang2@albany.edu,
xli48@albany.edu,
xi.wu@cuit.edu.cn,
xwang56@albany.edu}}
}

\begin{document}

\maketitle

\begin{abstract}
    Arbitrary style transfer aims to apply the style of any given artistic image to another content image. Still, existing deep learning-based methods often require significant computational costs to generate diverse stylized results. 
Motivated by this, we propose a novel reinforcement learning-based framework for arbitrary style transfer \textit{RLMiniStyler}. 
This framework leverages a unified reinforcement learning policy to iteratively guide the style transfer process by exploring and exploiting stylization feedback, generating smooth sequences of stylized results while achieving model lightweight.  
Furthermore, we introduce an uncertainty-aware multi-task learning strategy that automatically adjusts loss weights to adapt to the content and style balance requirements at different training stages, thereby accelerating model convergence. 
Through a series of experiments across image various resolutions, we have validated the advantages of \textit{RLMiniStyler} over other state-of-the-art methods in generating high-quality, diverse artistic image sequences at a lower cost. Codes are available at \url{https://github.com/fengxiaoming520/RLMiniStyler}.
\end{abstract}

\begin{figure*}[t]
\centerline{
  \includegraphics[width=1\textwidth]{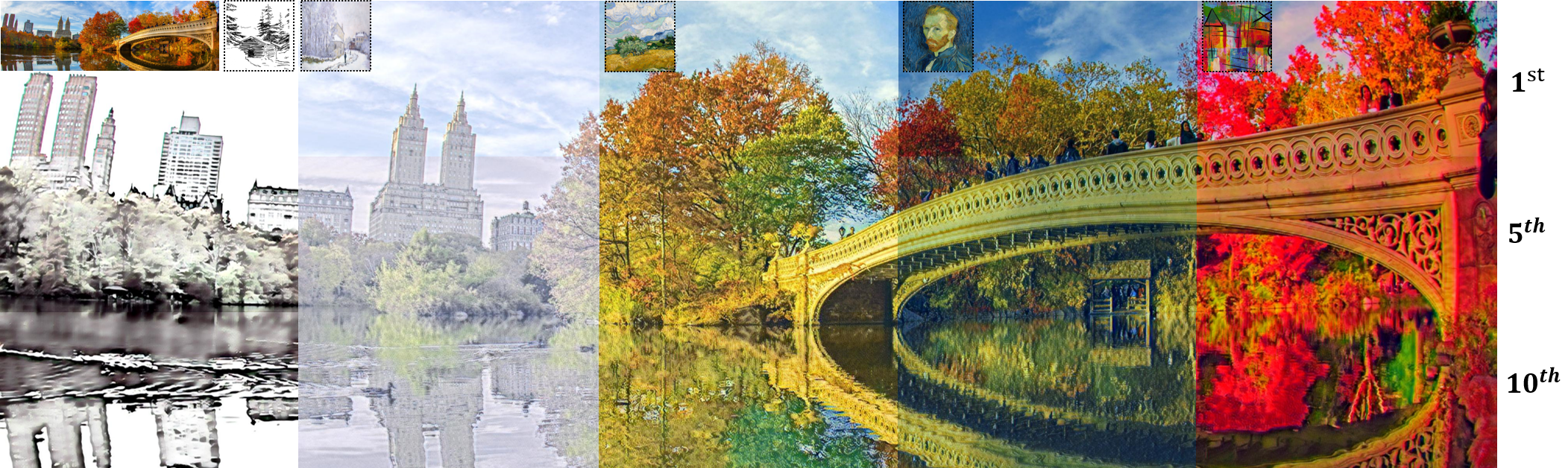}
  \vspace{-2mm}  
}
\caption{ \small
Illustration of our arbitrary style sequence generation process. 
\textbf{Top Left:} Content and Style Images (5 style examples).
\textbf{Right:} The sequence number of the results.
Content images are progressively stylized with increasing strength along prediction sequences (see the index). 
Our method allows for easy control over stylization degree, preserving content details in early sequences and synthesizing more style patterns in later sequences, resulting in a user-friendly approach. 
}
\label{introdemo} 
\vspace{-3mm}
\end{figure*}

\section{Introduction}

\label{sec:introduction}

The goal of style transfer is to alter the style of an image while preserving its content. Arbitrary style transfer (AST), a key task in this domain, involves the challenge of using a single model to apply any desired artistic style to any given content.
Since the pioneering work of Gatys et al.\cite{gatys2015neural} in neural style transfer, subsequent research\cite{an2021artflow,wu2022completeness,deng2022stytr2,kwon2023aesfa,lin2021drafting,liu2021adaattn,park2019arbitrary,wang2023microast} has made significant strides in enhancing model generalization capabilities, optimizing result quality, and accelerating inference speeds. 
Due to the varying preferences for the degree of stylization among individuals, precisely controlling the level of stylization to meet diverse needs is a challenging task.
Mainstream approaches typically rely on manual tuning of hyperparameters 
to balance content and style, achieving results with varying degrees of stylization
~\cite{gatys2015neural,huang2017arbitrary,liu2021adaattn,park2019arbitrary}, 
including the mixing ratio of content features to style features, as well as the individual weightings of content loss and style loss.
However,
repetitive process of trial and adjustment for achieving suitable weighting parameters, 
along with the complexity of networks with over 7 million parameters, limits their applicability. To simplify network models, MicroAST~\cite{wang2023microast} employs a streamlined model without pre-trained networks for faster inference, and AesFA~\cite{kwon2023aesfa} decomposes images into frequency components for efficient stylization. 
Even though these lightweight AST methods ensure computational efficiency and can perform style transfer on any style, 
{they also require manual adjustment of hyperparameters and retraining to achieve varying degrees of stylization for specific styles. Importantly, achieving a good balance between content and style through manual hyperparameter tuning is challenging and often results in under-stylization or over-stylization.}
Hence, it is necessary to develop a new arbitrary style transfer technique that not only facilitates the transfer of any style but also offers a rich array of style degree options for each particular style, relies less on manual hyperparameter tuning, and remains computationally efficient. Recently, RL-NST~\cite{feng2023controlling} pioneered the application of reinforcement learning to the single style transfer task, achieving precise control over the degree of stylization for one specific style. 
But it struggles to distinguish between diverse styles, requiring retraining when faced with new styles.

This paper proposed a novel framework named RLMiniStyler that leverages reinforcement learning for controlling the process of the arbitrary style transfer using a unified policy and uncertainty-aware automatic multi-task learning.  
Leveraging the autonomous exploration inherent in reinforcement learning, our proposed method refines style expression, resulting in a diverse range of stylized results.
By integrating a unified policy capable of effectively encoding both content and style images within a single neural network without feature confusion,
RLMiniStyler can use one encoder to extract content and style features, thereby reducing model complexity and ensuring a consistent approach to learning and adaptation.
{Compared to using two encoders, this design 
is more conducive to stable training in the reinforcement learning process.
} 
Additionally, the uncertainty-aware automatic multi-task learning allows for dynamic adjustment of learning priorities based on the current performance state.
Capable of rapidly generating a diverse array of results with varying degrees of stylization under limited resources, our method offers a richer visual experience beyond a singular result, as shown in Fig.~\ref{introdemo}. 

RLMiniStyler empowers the agent to autonomously learn and explore various style transformation strategies without being constrained by pre-defined rules, resulting in more diverse and innovative stylized images. 
In summary, we summarize the main contributions of this work as follows:

\begin{itemize}
\item We present the first method of arbitrary style transfer based on reinforcement learning.  RLMiniStyler provides a stable and flexible control of stylization. It allows flexible control over the degree of stylization by progressively incorporating style patterns into the results over time. 
    
\item We propose a unified policy within RLMiniStyler to ensure it remains sufficiently lightweight to operate efficiently in resource-constrained environments, while still maintaining high performance. 


\item We propose an uncertainty-aware, multi-task learning optimization strategy within our RLMiniStyler to automatically balance style learning and content preservation.
    
\item  Through comprehensive experiments on diverse image resolutions, we show the effectiveness of RLMiniStyler in creating high-quality and varied artistic sequences, showcasing its lightweight model advantage and superior or comparable performance across various evaluation metrics relative to both existing lightweight and state-of-the-art style transfer methods.
\end{itemize}

\section{Related Work}
\label{sec:related_work}


\textbf{Arbitrary Style Transfer (AST).} 
AST aims to enable style transfer using a single trained model, achieving a balance between content and style across various style images without requiring additional training. 
While recent advancements~\cite{deng2022stytr2,gu2018arbitrary,hu2023attention,huang2017arbitrary,liu2021adaattn,park2019arbitrary,wang2020diversified} have been made in this area, many methods have complex models and offer limited diversity in stylization results. 
Although recently proposed lightweight methods~\cite{wang2023microast,kwon2023aesfa} have employed lightweight models, 
they necessitate retraining to realize results with varying degrees of stylization for a particular style.
{Using pruning techniques~\cite{wu2024lighting} can also achieve lightweight style transfer models, 
but this approach inevitably leads to a decline in style transfer performance, such as insufficient stylization.
}



\textbf{Deep Reinforcement Learning for Neural Style Transfer.} 
The agent in reinforcement learning (RL) focuses on developing optimal strategies through continual exploration and exploitation
to maximize cumulative rewards. Handling high-dimensional continuous state and action spaces is particularly challenging for RL agents. 
Maximum Entropy Reinforcement Learning (MERL) methods~\cite{haarnoja2017reinforcement,haarnoja2018soft,hu2023attention,zhao2019uncertainty} demonstrate robust performance in high-dimensional continuous RL tasks by encouraging exploration. However, they may face limitations when applied to generative tasks such as Image-to-Image Translation (I2IT), as they are not inherently designed for generative models. SAEC~\cite{luo2021stochastic}, a framework that extends the traditional MERL approach, introduces a generative component to effectively handle I2IT tasks, but the 1D action space limits its effectiveness when attempting to process images with resolution higher than $128\times128$.
Recently, RL-NST~\cite{feng2023controlling} has successfully extended SAEC 
to the style transfer task by expanding the action space to 2D and 3D. However, as a single-style transfer method, it requires retraining for each new style, making it unsuitable for the AST task. 


\begin{figure*}[t]
\centerline{
  \includegraphics[width=0.95\textwidth]{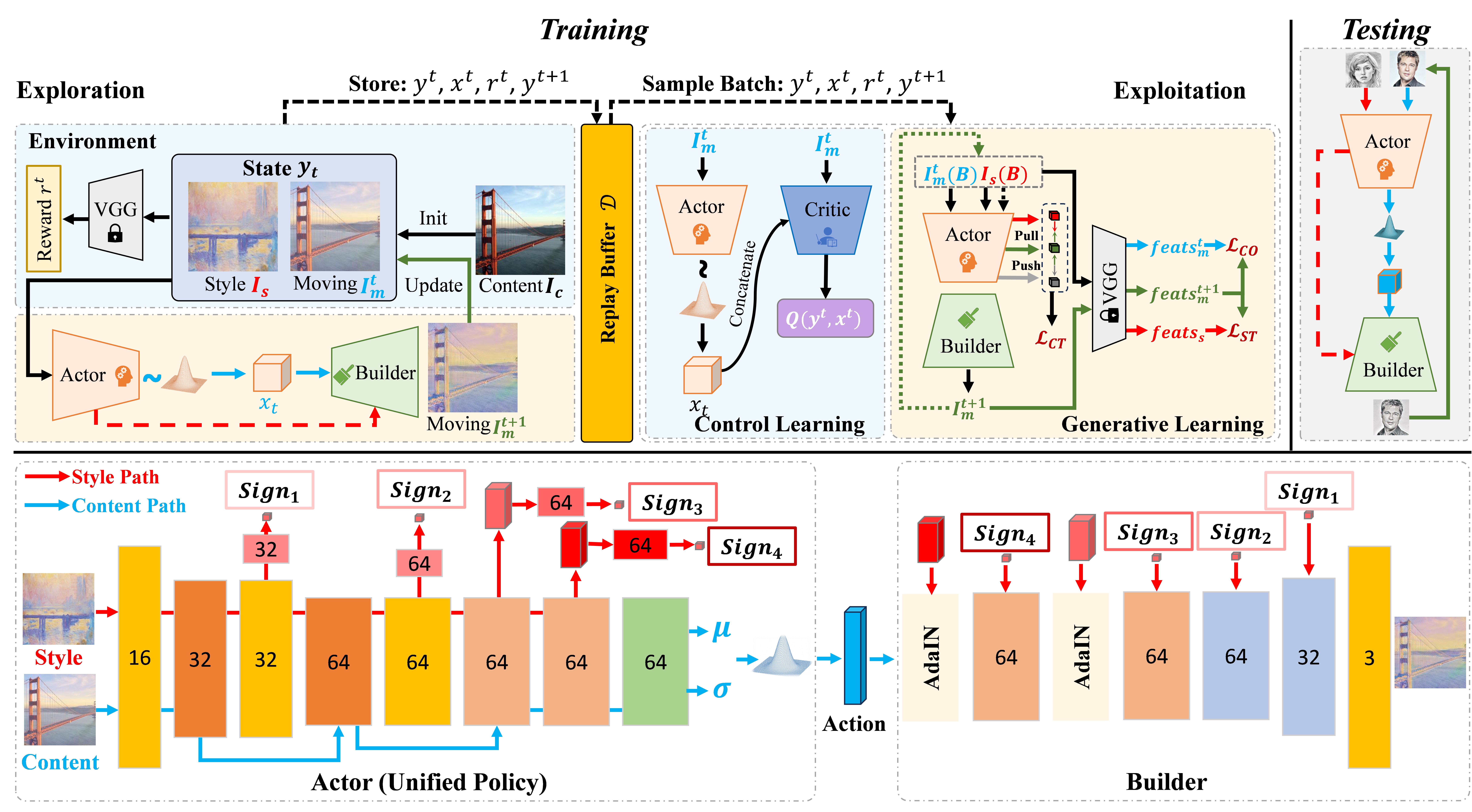}
}
\caption{Overview of the RLMiniStyler model. 
\textbf{Top:}
The state $y_t$ is initialized with the content image $I_c$ and the style image $I_s$. Latent-action ${\bf x}_t$ is sampled from a high-dimensional Gaussian distribution and is concatenated with the critic's output. It is estimated by the policy $P_\kappa$: ${\bf x_t} \sim P_\kappa({\bf x}_t|{\bf y}_t)$. The predicted moving image $I_m^{t+1}$ is generated by builder $B_\tau$. 
'Pull' and 'Push' refer to minimize and maximize the distance between two feature maps, respectively.
\textit{Note that the pre-trained VGG network is used only to extract features for calculating rewards and losses during training.}
\textbf{Bottom:} The structure of the actor and the builder. And $Sign_{1,2,3,4}$ refer to the style signals derived from the calculation of style features. Different colors in the network represent different network architectures, and
details of the network structure can be found in supplementary materials. 
}
\label{fig:pipline}
\end{figure*}

\begin{figure*}[t]
    \centering
    \includegraphics[width=1\textwidth]{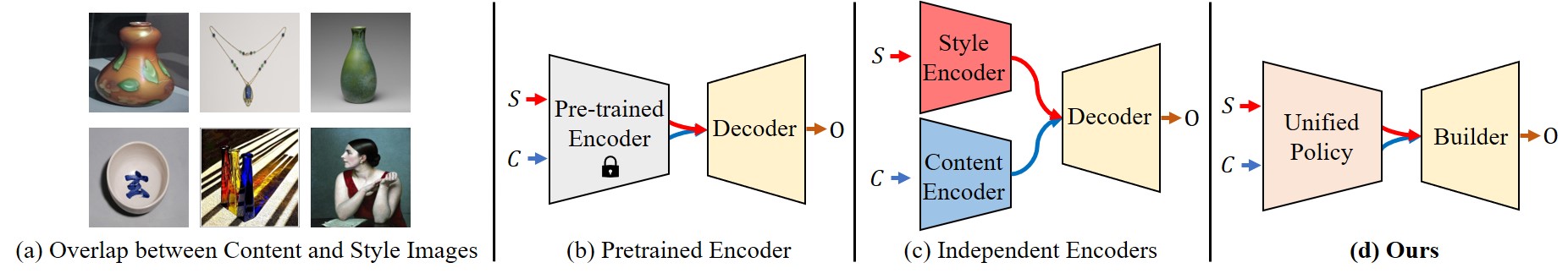}
    \caption{
    The overlap between style and content images, and the illustration from Pre-trained Encoder to Non-pretrained Encoder. 
    In the figure, 'S', 'C', and 'O' represent the style image, content image, and stylized output, respectively.
    The images shown in (a) are drawn from both content and style datasets, but the boundaries between them are so blurred that it's challenging to clearly distinguish their original sources.
    Most existinig style transfer methods typically employ two encoding approaches: one directly utilizes single complex pre-trained encoder (b), while the other trains separate encoders for content and style (c). In contrast, our method adopts a novel approach, using single mini-unified policy for both content and style (d). We detail the specifics of this unified policy as shown in Fig.~\ref{fig:pipline}. 
    }
    \label{fig:encoder_compare}
\end{figure*}

\section{Method}
\label{sec:Method}
Existing AST methods usually use complex neural networks for one-step inference, limiting stylization diversity and restricting user preferences. Based on MERL framework~\cite{haarnoja2018soft}, we propose a novel, lightweight RL method for AST to enhance the richness of artistic stylization. In our method, style transfer is regarded as a sequential decision-making problem. In the guidance of a well-defined reward function, our RL agent selects optimal actions at each time step, and generates intermediate stylized results with varying style degrees accordingly.
The overview of our method is shown in Fig.~\ref{fig:pipline}. 
Our approach includes three key components: the actor ${P}_\kappa$ characterized by parameters $\kappa$, the builder ${B}_\tau$ characterized by parameters $\tau$, and the critic ${Q}_\delta$ characterized by parameters $\delta$. 
The actor serves as the unified policy network responsible for making decisions and style feature extraction 
based on the current state composed of both the moving image and the style image,
the builder acts as the generation network responsible for executing the actor's stylized decisions, 
and the critic acts as the scoring network responsible for evaluating the actor's decisions. 
The actor and the critic constitute the RL learning path for style control, while the actor and the builder constitute the generative learning part for generating stylized image.
We next describe our method in details.




\subsection{Deep Reinforcement NST Framework}

In our RL environment $\Upsilon$, $C_D$ and $S_D$ represent the content dataset and the style dataset, respectively. 
The \textbf{state} ${\bf y}^t \in \Upsilon$ is composed of two parts: the moving image $I_m^t$ and the style image $I_s \in S_D$. The moving image $I_m^t$ is initialized using the content image $I_c \in C_D$. 
The \textbf{action} ${\bf x}^t$ is determined by the agent based on its observation of the current state. To extract high-level abstract actions from the actor, we establish a model of stochastic latent actions conditioned on the current state (that is, the action ${\bf x}^t$ follows the conditional probability ${\bf x}^t = {P}_{\kappa}({\bf x}^t|{\bf y}^t)$). In practice, we employ the reparameterization technique~\cite{kingma2013auto} to obtain these actions.
The moving image $I_m^{t+1}$ in state ${\bf y}^{t+1}$
is created by the builder based on ${\bf x}^t$ and current state ${\bf y}^t$.
The \textbf{reward} ${\bf r}^t$ is derived from the measurement of style discrepancy between the moving image $I_m^t$ and the style image $I_s$. The reward is inversely proportional to this discrepancy, such that a smaller style difference results in a larger reward.


\subsection{Unified Policy for Efficient Style and Content Representation}

Existing AST models~\cite{deng2022stytr2,huang2017arbitrary,johnson2016perceptual,kwon2023aesfa,liu2021adaattn,park2019arbitrary,wang2023microast} widely employ an encoder-decoder network as backbone architecture. Most of them ~\cite{gu2023two,huang2017arbitrary,li2017universal,liu2021adaattn,park2019arbitrary} use pre-trained VGG ~\cite{simonyan2014very} as the encoder due to its strong capability to capture a wide range of features useful for representing both content and style in images, as shown in Fig.~\ref{fig:encoder_compare}(b). But the the complexity of the pre-trained VGG can lead to substantial computational expenses and may introduce unwanted style patterns, like "eyes".
An alternative approach~\cite{deng2022stytr2,wang2023microast,kwon2023aesfa} involves training two  encoders to independently process content and style images, treating them as separate distributions, as shown in Fig.~\ref{fig:encoder_compare}(c). In this way, more appropriate encoding of content and style images is achieved to avoid incorrect style patterns. However, as shown in Fig.~\ref{fig:encoder_compare}(a), there are no clear boundaries between content images and style images in practice. 
Hence, due to the inherent overlap between these two types of images being overlooked, using two separate encoders complicates the AST model and slows down the training process.

In light of this, RLMiniStyler leverages a unified policy for modeling content and encoding style using a single encoder, as shown in Fig.~\ref{fig:encoder_compare}(d). To enable the encoder to have the ability to simultaneously process both content and style images,
we draw inspiration from the StyleBank ~\cite{chen2017stylebank}, which decouples content and style images through explicit style representation. Specifically, we integrated two additional style space dedicated to style encoding at different positions within the encoder's architecture, while the other parts of the encoder were used for general feature extraction, as shown in Fig.~\ref{fig:pipline}. 
This design not only maintains the efficiency of a single encoder but also enhances our control over the subtleties between content and style images by processing style features at different levels. 
Compared to the design of two separate encoders, the actor, capable of perceiving content and style simultaneously, can make more precise decisions based on the current state.
In other words, we can more accurately manipulate the outcome of style transfer to achieve a richer variety of stylistic fusion effects.

\subsection{Joint Learning}

Our framework employs a joint learning strategy, integrating two mutually coordinated optimization processes: control learning and generative learning.
In the  control learning, our model learns control policies, while in the generative learning, it learns stylized image generation. 
Training alternates between these two parts. 
The generative learning consists of the Actor  $P_{\kappa}$ and the Builder $B_\tau$, and the design of the loss function helps in effectively propagating gradient information between the Actor $P_{\kappa}$ and the Builder $B_\tau$. 
The  control learning consists of the Actor $P_{\kappa}$ and the Critic $Q_{\delta}$, and the training of the Actor can be conducted jointly through the  control learning and the generative learning to ensure rapid and stable convergence. 
Algorithm in appendix describes the RLMiniStyler algorithm. 
All parameters are optimized based on the samples from replay pool $\cal D$.


\subsubsection{ Control Learning}




In the  control learning, 
adhering to the MERL framework~\cite{haarnoja2018soft}, we iteratively refine a stochastic policy $P_\kappa$ utilizing reward signals ${\bf r}^t$ and soft Q-values $Q_{\delta}({\bf y}^t, {\bf x}^t)$. Here, the action ${\bf x}^t$ is generated by the actor 
in response to the current state ${\bf y}^t$, following the policy $P_{\kappa}({\bf x}^t|{\bf y}^t)$. The soft Q-function $Q_{\delta}({\bf y}^t, {\bf x}^t)$, computed by the critic network, provides an estimation of the expected cumulative reward for the state-action pair $({\bf y}^t, {\bf x}^t)$ under the current policy. During the evaluation phase, we guide the improvement of the stochastic policy through the minimization of the soft Bellman residual, defined as:
%
\begin{equation}
    \begin{aligned}
    J_Q(\delta) =& 
    \mathbb{E}_{({\bf y}^t, {\bf x}^t, r^t, {\bf y}^{t+1})\sim  \mathcal{D}} \big[\frac{1}{2} \Big(Q_{\delta}({\bf y}^t, {\bf x}^t) \\&-  \big(r^t + \gamma \mathbb{E}_{{\bf y}^{t+1}}\left[V_{\bar{\delta}}({\bf y}^{t+1})\right]\big)\Big)^2 \big],
    \end{aligned}
\end{equation}
where $\mathcal{D}$ is the replay pool and $V_{\bar{\delta}}({\bf y}^{t}) = \mathbb{E}_{{\bf x}^{t}\sim P_\kappa}[Q_{\bar{\delta}}({\bf y}^{t}, {\bf x}^{t}) - \alpha \log P_\kappa ({\bf x}^{t}|{\bf y}^{t})]$. 
We set the reward signal ${\bf r}^t$ as the negative of the style loss 
to ensure that the agent learns to control the stylization process. Given our objective is style-related, such a choice of reward function is reasonable. Specifically, the style loss serves as a simple and effective means to assess the similarity between the stylized output and the target style image, hence we set the reward function as the negative of the style loss ${\bf r}^t=-\mathcal{L}_{ST}$, where the detailed definition of $\mathcal{L}_{ST}$ is shown in Eq.~\eqref{eq:style_loss}.



The target critic network, denoted as $Q_{\bar{\delta}}$, plays a crucial role in stabilizing the training process. The parameters of this network, represented as $\bar{\delta}$, are determined by calculating the exponential moving average of the parameters from the critic network~\cite{lillicrap2015continuous}: $\bar{\delta} \rightarrow \omega \delta + (1-\omega)\bar{\delta}$, with hyperparameter $\omega\in [0,1]$. To optimize $J_Q(\delta)$,  we use the gradient descent with respect to parameters $\delta$ as:
\begin{equation}
    \begin{aligned}
    \delta \leftarrow & \delta - \rho_Q \triangledown_{\delta} Q_{\delta}({\bf y}^t, {\bf x}^t)\Big(Q_{\delta}({\bf y}^t, {\bf x}^t) - r^t \\
    &- \gamma \left[Q_{\bar{\delta}}({\bf y}^{t+1}, {\bf x}^{t+1}) - \alpha \log P_{\kappa} ({\bf x}^{t+1}|{\bf y}^{t+1})\right]\Big),
    \end{aligned}
\label{eq:update_delta}
\end{equation}
where $\rho_Q$ is the learning rate. 
In the RL framework, the critic evaluates the actions taken by the actor, which in turn influences the policy decisions of the actor. Consequently, the following objective can be applied to minimize the Kullback-Leibler (KL) divergence between the policy induced by the actor and a Boltzmann distribution, as determined by the Q-function:
\begin{equation}
    \begin{aligned}
    J_{P} (\kappa) =& \mathbb{E}_{ {\bf y}^t\sim  \mathcal{D}}\big[ \mathbb{E}_{{\bf x}^t\sim  P_{\kappa}} \left[\alpha \log (P_{\kappa}({\bf x}^t| {\bf y}^t))-Q_\delta({\bf y}^t,{\bf x}^t)\right]\big]\\
    =& \mathbb{E}_{ {\bf y}^t\sim  \mathcal{D}, {\bf n}^t\sim \mathcal{N} (\bm{\mu},\bm{\Sigma}) } 
    \big[\alpha \log (P_{\kappa}( f_P({\bf n}^t,{\bf y}^t)|{\bf y}^t))
    \\&- Q_\delta({\bf y}^t,f_P({\bf n}^t,{\bf y}^t))\big].
    \end{aligned}
\end{equation}
The last equation holds because ${\bf x}^t$ can be evaluated by $f_P({\bf n}^t,{\bf y}^t)$, where ${\bf n}^t$ is a noise vector sampled from a 3D Gaussian distribution with mean $\bm{\mu}=0$ and standard deviation $\bm{\Sigma}=1$.
Note that hyperparameter $\alpha$ can be automatically adjusted by using the method proposed in~\cite{haarnoja2018soft}. Similarly, we apply the gradient descent method with the learning rate $\rho_\kappa$ to optimize parameters as:
\begin{equation*}
    \begin{aligned}
    {\bf \kappa} \leftarrow & \kappa -\rho_\kappa\Big(\triangledown_\kappa \alpha \log(P_\kappa({\bf x}^t|{\bf y}^t)) + \big(\triangledown_{{\bf x}^t}\alpha \log(P_{\kappa}({\bf x}^t|{\bf y}^t))\!
    \\& -\!\triangledown_{{\bf x}^t}Q_{\delta}({\bf y}^t,{\bf x}^t)\big) \triangledown_\kappa f_\kappa ({\bf n}^t,{\bf y}^t)\Big).
    \end{aligned}
\label{eq:update_P}
\end{equation*}





\begin{figure*}[t]
\centerline{
  \includegraphics[width=1\textwidth]{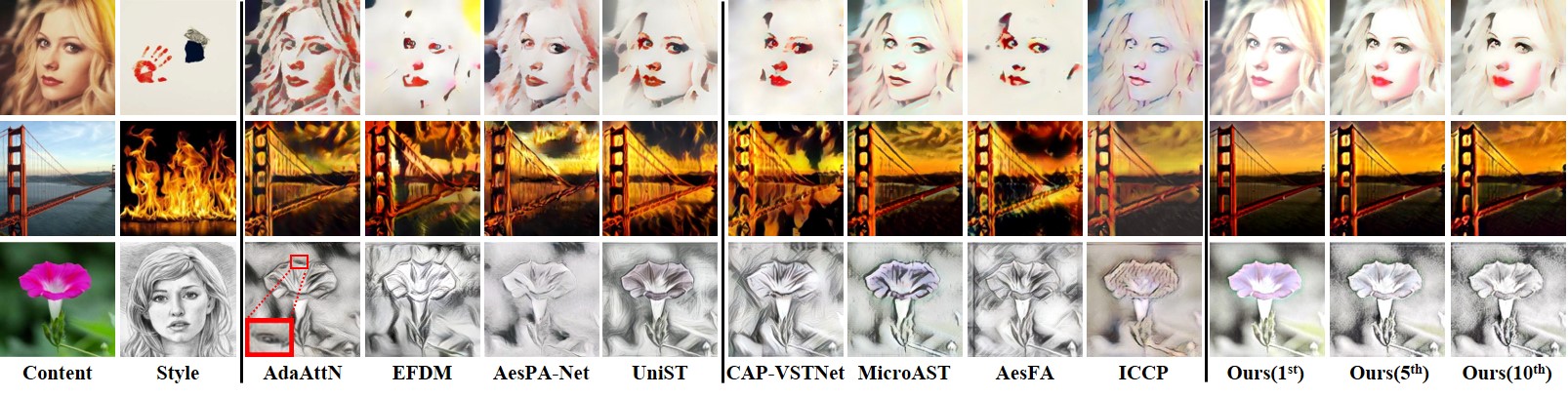}
}
\caption{Qualitative Comparison with several AST algorithms in 256 pixel resolution. The 1st and 2nd columns present the content and style images, respectively. The subsequent four columns display the results from the current SOTA AST methods. The three columns immediately following showcase the results of the lightweight methods. Lastly, we present the sequential stylization results generated by our method, including sequences 1st, 5th, and 10th.
}
\label{fig:vision_compare}
\end{figure*}

\subsubsection{Generative Learning} \label{sec:generative-learning}
Generative learning, through specific training strategies, enhances our model's generation capability in style transfer, ensuring the production of high-quality stylized image results.
We assess the similarity between the stylized image $I_m^{t+1}$ and the input images by comparing their high-level features. Specifically, we employ a pre-trained VGG~\cite{simonyan2014very} as a feature extraction backbone $\phi$ to extract features independently from the moving image $I_m^t$, the style image $I_s$, and the stylized image $I_m^{t+1}$. We calculate the content loss $\mathcal{L}_{CO}$ by comparing the semantic similarity  between $I_m^{t+1}$ and $I_m^{t}$ and the style loss $\mathcal{L}_{ST}$ by comparing the style similarity between $I_m^{t+1}$ and $I_s$. 

\textbf{Content Loss.}  
We evaluate how closely the stylized image resembles the content image, by maximizing perceptual similarity using the widely adopted perceptual loss~\cite{johnson2016perceptual}.
Let $\phi^{(j)}$ denote the activation of the $j$-th layer, producing a feature map with dimensions $C^j \times H^j \times W^j$, where $C^j$, $H^j$, and $W^j$ represent the number of channels, height, and width of the feature map, respectively. The content loss $\mathcal{L}_{CO}$ is calculated by:
\begin{equation}
\mathcal{L}_{CO}(I_m^{t+1},I_m^{t})=\frac{1}{C^jH^jW^j}\parallel \phi^{(j)}(I_m^{t+1})- \phi^{(j)}(I_m^{t})\parallel _{2}^{2}.
\end{equation}


\textbf{Style Loss.} 
The {style loss $\mathcal{L}_{ST}$} estimates the style deviations between the stylized image $ I_m^{t+1} $ and style image $ I_s $. Let $J$ represent the layer number of the network $\phi$.
It calculates statistical measures of $\mu$ and standard deviation $\sigma$ to penalize \( I_m^{t+1} \), inspired by \cite{huang2017arbitrary}:

\begin{equation}
\begin{aligned}
\mathcal{L}_{ST}(I_m^{t+1}, I_s) &= \sum_{j=1}^J{\parallel}\mu (\phi^{(j)}(I_m^{t+1}))-\mu (\phi^{(j)}(I_s))\parallel _2^2 \\
& +\sum_{j=1}^J{\parallel}\sigma (\phi^{(j)}(I_m^t))-\sigma (\phi^{(j)}(I_s))\parallel _2^2.
\end{aligned}
\label{eq:style_loss}
\end{equation}

\textbf{Hierarchical Style Representation Contrastive Loss (HSRCL).} Recent studies~\cite{chen2021artistic,wang2023microast} have demonstrated that 
the lightweight network struggles to fully capture and express the style features of style images in a single inference process,
and incorporating a contrastive learning loss can mitigate this issue. 
For instance, MicroAST~\cite{wang2023microast} employs a style signal contrastive learning loss to deal with this, 
but it predominantly relys on deep-layer features for contrastive learning, overlooking the contributions of shallow-layer features to the overall style representation.
To this end, we introduce a novel hierarchical style representation contrastive loss, which integrates contrastive learning between deep and shallow feature representations, so as to enhance the style representation. More specifically, 
when sampling a batch of data from the replay buffer $\cal D$, we construct both positive and negative sets for each sample's deep and shallow features.
And the feature contrastive
loss respectively computed from the deep features and shallow features are combined to create a hierarchical style contrastive loss function $\mathcal{L}_{CT}$, which is defined as:
%
\begin{equation}
    \mathcal{L}_{CT}=\sum_{i=1}^N{\sum_{k=1}^K{\frac{\parallel 
    P_\kappa(I_m)^{(i,k)} - P_\kappa(I_s)^{(i,k)}
    \parallel _2^2}{\sum_{j\ne i}^N{\parallel}
    P_\kappa(I_m)^{(i,k)} - P_\kappa(I_s)^{(j,k)}
    \parallel _2^2}}},
\end{equation}
where $K$ represents the number of feature layers in the unified policy network, $N$ represents the batch size. The batch comprises $N$ states ${\bf Y}=\{{\bf y}_1, {\bf y}_2, ..., {\bf y}_N\}$. Each state ${\bf y}_{i} \in {\bf Y}$ consists of a moving image $I_m$ and a style image $I_s$. 
For each $I_m$, we consider the style image $I_s$ from ${\bf y}_{i}$ as a positive sample and the style images $I_s$ from other ${\bf y}_{j }$ as negative samples, where $j\ne i$.

\begin{table*}[t]
\centerline{
\resizebox{1.0\linewidth}{!}{
\begin{tabular}{c|cc|ccclc|ccclc}
\hline
         & \multicolumn{2}{c|}{Model Complexity}      & \multicolumn{5}{c|}{256$\times$256 Pixel Resolution}         & \multicolumn{5}{c}{512$\times$512 Pixel Resolution}                       \\ \hline
Method        & \multicolumn{1}{c}{Params (1e6) $\downarrow$} & \multicolumn{1}{c|}{Storage (MB) $\downarrow$} & Content Loss 
 $\downarrow$    & SSIM $\uparrow $          & \multicolumn{1}{c|}{Style Loss $\downarrow$}      & \multicolumn{1}{c|}{Time(s) $\downarrow$} & Pref.(\%) $\uparrow$ & Content Loss $\downarrow$ & SSIM $\uparrow$ & \multicolumn{1}{c|}{Style Loss $\downarrow$} & \multicolumn{1}{c|}{Time(s) $\downarrow$} & Pref.(\%) $\uparrow$ \\ \hline
AdaAttN(2021)     &13.6299   &128.4020  &3.0668   &0.4987   &\multicolumn{1}{c|}{0.6027} &\multicolumn{1}{c|}{0.0117} &12.00   &2.4280   &0.5341  &\multicolumn{1}{c|}{0.5516}  &\multicolumn{1}{c|}{0.1032}  &10.67  \\
EFDM(2022)        &7.0110    &26.7000  &3.6671   &0.3165   &\multicolumn{1}{c|}{0.4233} &\multicolumn{1}{c|}{0.0073}  &4.67 &2.9439   &0.3788   &\multicolumn{1}{c|}{0.3268}  &\multicolumn{1}{c|}{0.0079} &6.67    \\
CAP-VSTNet(2023)  &4.0899    &15.6719  &3.5984   &0.4501   &\multicolumn{1}{c|}{\textbf{0.3151}} &\multicolumn{1}{c|}{0.0423} &7.33   &2.7459   &0.4864   &\multicolumn{1}{c|}{\textbf{0.2234}}  &\multicolumn{1}{c|}{0.1209} & 7.33 
 \\
AesPA-Net(2023)   &23.6737   &92.3340  &2.6822   &0.4504   &\multicolumn{1}{c|}{0.8266}  &\multicolumn{1}{c|}{0.3110} &5.67  &2.0412   &0.5195   &\multicolumn{1}{c|}{0.8756}  &\multicolumn{1}{c|}{0.4628}  &10.00   \\
UniST(2023)       &65.2545   &302.9424  &2.8888   &0.4305   &\multicolumn{1}{c|}{0.4137} &\multicolumn{1}{c|}{0.0295}  &9.67   &2.4080   &0.4567   &\multicolumn{1}{c|}{0.2952}  &\multicolumn{1}{c|}{0.0347} &7.33   \\
MicroAST(2023)    &0.4720    &1.8570  &2.6382   &0.4753   &\multicolumn{1}{c|}{0.6247}  &\multicolumn{1}{c|}{\textbf{0.0066}}  &9.00  &2.0349   &0.5034   &\multicolumn{1}{c|}{0.4960}  &\multicolumn{1}{c|}{\textbf{0.0069}} &7.67  
 \\
AesFA(2024)       &3.2208    &12.3100  &3.3734   &0.4115   &\multicolumn{1}{c|}{0.3945}  &\multicolumn{1}{c|}{0.0167}  &8.33  &2.7624   &0.4466   &\multicolumn{1}{c|}{0.3024}  &\multicolumn{1}{c|}{0.0187} &7.00   \\
ICCP(2024)       &\textbf{0.0790}    &\textbf{0.3447}  &{2.7964}   &0.5152   &\multicolumn{1}{c|}{1.3025}  &\multicolumn{1}{c|}{0.0087}  &3.33  &2.1236   &0.5559   &\multicolumn{1}{c|}{1.1415}  &\multicolumn{1}{c|}{0.0098} &4.00   \\ \hline
Ours($1^{st}$)      &0.3712    &{1.4750}  &\textbf{1.1684}   &\textbf{0.6444}   &\multicolumn{1}{c|}{1.0487} &\multicolumn{1}{c|}{0.0094}  &\textbf{15.00}   &\textbf{0.9292}   &\textbf{0.6517}   &\multicolumn{1}{c|}{0.8927}  &\multicolumn{1}{c|}{0.0150}   &\textbf{17.33}   \\
Ours($5^{th}$)      &0.3712    &{1.4750}  &\textbf{2.1508}   &\textbf{0.5509}   &\multicolumn{1}{c|}{0.6974} &\multicolumn{1}{c|}{0.0336}  &\textbf{20.33}   &\textbf{1.6491}   &\textbf{0.5711}   &\multicolumn{1}{c|}{0.5528}  &\multicolumn{1}{c|}{0.0852}  &\textbf{13.67}  \\
Ours(${10}^{th}$)     &0.3712    &{1.4750}  &2.7518   &0.4898   &\multicolumn{1}{c|}{0.6209} &\multicolumn{1}{c|}{0.0631}   &4.67   &2.0871   &0.5191   &\multicolumn{1}{c|}{0.4892}  &\multicolumn{1}{c|}{0.1733}  &8.33   \\ \hline
\end{tabular}}
}
\caption{Quantitative Comparison of Model Complexity and Performance with Various AST Algorithms at Standard Resolutions. `Pref.' represents user preferences from our user study.  }
\label{tab:compare_256_512}
\end{table*}

\textbf{Uncertainty-aware Automatic Multi-task Learning.} 
A common way to enhance the quality of style transfer involves quantifying the semantic similarity to the content image and the style similarity to the style image through content and style loss functions,
along with auxiliary loss functions, such as adversarial loss. 
But the weights for these loss functions are usually heuristically selected before training and remain unchanged throughout the training process, which is not sufficient enough to handle  images with different style and content. 



To this end, as inspired by \cite{kendall2018multi}, we propose to use a multi-task learning framework that treats content learning, style learning, and contrastive learning as distinct but interconnected tasks. Using {\em homoscedastic uncertainty}, we dynamically adjust the loss weights of each task derived from a principled probabilistic model, achieving a balanced optimization objective that adapts throughout training. Unlike traditional methods requiring manual tuning of loss weights, our approach learns the relative importance of each task's loss function directly from the data. 
This not only simplifies the training process but also enables the dynamic modulation of the content loss and the style loss ratios to find the optimal solution.

Let $\lambda_c$, $\lambda_s$, $\lambda_{ct}$ denote the loss weights for content loss, style loss, and contrastive loss, respectively. These weights can adapt based on homoscedastic uncertainty $\sigma_1^2$, $\sigma_2^2$, and $\sigma_3^2$, reflecting the noise level or task confidence. 
And they are inversely proportional to the noise parameters.
The final loss is:
\begin{equation}
\begin{aligned}
\mathcal{L}_{final}&(\kappa, \tau, \lambda_c, \lambda_s, \lambda_{ct}) = 
\lambda_c \mathcal{L}_{CO} + 
\lambda_s \mathcal{L}_{ST} + \lambda_{ct} \mathcal{L}_{CT} + \epsilon, \\
&\lambda_c=\frac{1}{\sigma_1^2}, \lambda_s=\frac{1}{\sigma_2^2}, \lambda_{ct}=\frac{1}{\sigma_3^2}, \epsilon = \log(\sigma_1 \sigma_2 \sigma_3),
\end{aligned}
\label{eq:awl}
\end{equation}
where $\log(\sigma_1 \sigma_2 \sigma_3)$ acts as a regularizer to prevent excessive increase in noise.  
Lastly, we employ a gradient descent method with the learning rate $\eta$ to update the Actor and Builder parameters ($\kappa$ and $\tau$) as well as 
$\sigma_i(i= 1, 2, 3)$:
%
\begin{equation}
    \kappa \leftarrow \kappa - \eta_\kappa \triangledown_\kappa \mathcal{L}_{final}, \quad
    \tau \leftarrow \tau - \eta_\tau \triangledown_\tau \mathcal{L}_{final},
    \label{eq:update_dl}
\end{equation}
\begin{equation}
    \sigma_i \leftarrow \sigma_i - \eta_{\sigma_i} \triangledown_{\sigma_i} \mathcal{L}_{final}.
    \label{eq:update_dl_sigma}
\end{equation}



    


    

    

\section{Experiments}
\label{Experiment}


\subsection{Experimental Setup}
\label{sec:exp:setup}

{\bf Datasets and evaluation metric:}
Like most AST methods~\cite{deng2022stytr2,huang2017arbitrary,liu2021adaattn,park2019arbitrary,wang2023microast}, we utilize the MS-COCO dataset~\cite{lin2014microsoft} for content and the WikiArt dataset~\cite{phillips2011wiki} for style. During training, images are first scaled to 512$\times$512 pixels, then randomly cropped to 256$\times$256, while testing can handle any input size. Following MicroAST~\cite{wang2023microast}, we assess all algorithms across seven aspects: {\em visual effect, inference time, parameter count, content loss, style loss, SSIM~\cite{wang2004image}, and storage space}.

{\bf Implementation details:}
We use the Adam optimizer~\cite{kingma2014adam} with a learning rate 2e-4, the batch size in the environment set to 1, and the batch size sampled from the replay buffer set to 8. All experiments are conducted on a single NVIDIA Tesla P100 (16GB) GPU.

\subsection{Comparisons with Prior Arts}
\label{sec:exp:comparison}

{\bf Baselines:}
We compare our method with four light-weight AST methods: CAP-VSTNet~\cite{wen2023cap}, MicroAST~\cite{wang2023microast}, AesFA~\cite{kwon2023aesfa}, and ICCP~\cite{wu2024lighting}, as well as four state-of-the-art AST methods: AdaAttN~\cite{liu2021adaattn}, EFDM~\cite{zhang2022exact}, AesPA-Net~\cite{hong2023aespa} and UniST~\cite{gu2023two}.  All codes used in the experiment are sourced from their respective public repositories, and we use the default settings provided.

{\bf Qualitative comparison:}
We visually compare our method with all baseline methods in Fig.~\ref{fig:vision_compare}. AdaAttN shows a repetitive style pattern resembling the eyes (the third row), while EFDM and CAP-VSTNet lose a significant semantic and structural content (first and second rows). AesPA-Net produces inconsistent results, especially in the eye area (first row). UniST, MicroAST and ICCP show insufficient stylization (third row), and AesFA has severe boundary artifacts (third row). 
In contrast, our approach generates a sequence of results with increasing stylization levels while maintaining coherent content structure. 
{Our method has also been compared with lightweight baselines at higher resolutions (512, 4K). Due to space constraints, the detailed comparison results are included in the supplementary materials.}

{\bf Quantitative comparison:}
Table~\ref{tab:compare_256_512} provides a comprehensive comparison between our approach and baseline models.
Our method consistently achieves competitive scores in content loss, SSIM, style loss, and inference time, demonstrating its efficiency and effectiveness in producing outputs that balance style expression with content preservation. As the sequence progresses, our method enhances style richness while maintaining content fidelity. 
In terms of model complexity, our model outperforms the minimally pruned model in performance and features a lower parameter count and reduced complexity compared to the smallest non-pruned model.
{Similarly, more comparative results with lightweight methods at high resolutions (1K, 2K, 4K) are included in the supplementary materials.
}
Additionally, it is the first AST method capable of automatically controlling the degree of stylization on images ranging from 256 to 4K resolution.
\vspace{-0.3em}
\subsection{Ablation Study}
\label{sec:ablation}

\begin{figure}[t]
\centerline{
  \includegraphics[width=0.48\textwidth]{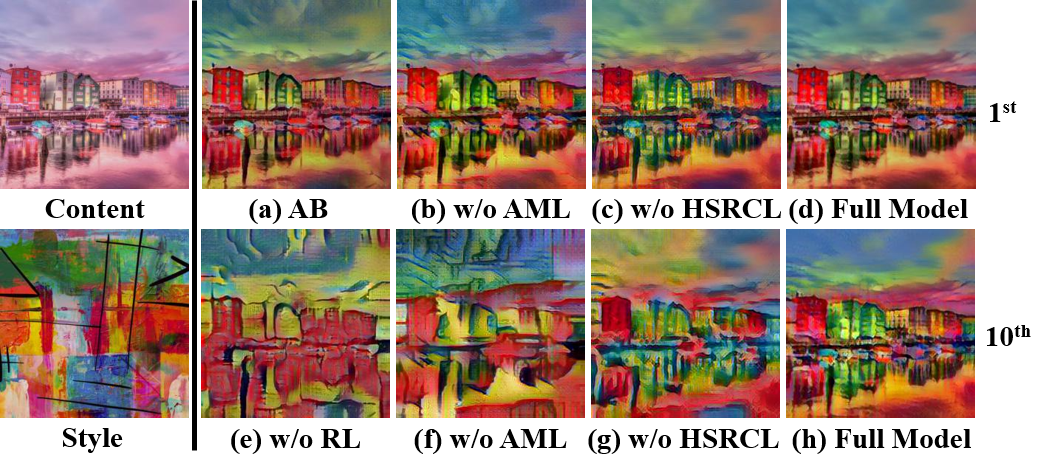}
}
\caption{
Ablation Study Results Comparing the Impact of RL, Automatic Multi-task Learning (AML), and 
Hierarchical Style Representation Contrastive Loss 
(HSRCL) vs. Style Signal Contrastive Loss on Style Transfer Performance. 
The visual comparison underscores the contributions of RL, AML, and HSRCL to the fidelity and stability of stylized results across sequences.
More results are presented in the supplementary materials.
}
\label{fig:ablationstudy1}
\end{figure}

\textbf{With and without RL:}
We discussed the effectiveness of RL in style control.
In Fig.~\ref{fig:ablationstudy1},
without RL, the Actor-Builder (AB) in (a) initially preserves semantic information. 
But as shown in (e), at sequence 10, notable content information is lost. In contrast, our method in (d) produces smoother and clearer stylized images from the start, and stably maintains high-quality results throughout the sequence in (h) at sequence 10. This consistent performance highlights the significant enhancement of RL provides to DL-based AST models.


\textbf{Automatic multi-task learning (AML) vs. manual settings:}
We manually tuned the loss weights in our method, based on the settings of MicroAST and empirical adjustments. Specifically, we set the content loss weight $\lambda_{c}=1$, the style loss weight $\lambda_{s}=3$, and the HSRCL loss weight $\lambda_{ct}=3$, while keeping all other settings unchanged. 
As shown in Fig.~\ref{fig:ablationstudy1} , compared to the fixed loss weight method in (b,f), our approach using AML demonstrates superior content preservation in both sequence 1 in (d) and sequence 10 in (h). Our study indicates that AML significantly enhances model performance and accelerates network convergence.

\textbf{Hierarchical style representation contrastive loss (HSRCL) vs. style signal contrastive loss:} 
We investigated the effectiveness of HSRCL by comparing with the deep-feature based contrastive loss proposed in MicroAST~\cite{wang2023microast}. 
As shown in Fig.~\ref{fig:ablationstudy1}, for sequence 1, using only deep features for contrastive learning (c) exhibits less of style diversity as compared with the result in (d). Comparing with sequence 10 in (h), there is a noticeable decline in (g) in terms of content affinity due to incoherent style expression. This experiment demonstrates that HSRCL significantly enhances the model's capacity in style expression.

\subsection{User Study}
\label{sec:user}

We conducted user study on nine different methods. We recruited 30 participants representing a diverse range of ages, genders, and professional backgrounds. Each participant was randomly presented with 20 ballots: 10 at a $256\times 256$ resolution and $512 \times 512 $ resolution. Each ballot included the content image, the style image, and 11 randomly shuffled stylized results. Note that since our method produces sequential results, we present the outcomes at the first, fifth, and tenth sequences. 
We collected 300 valid ballots for each resolution, and the detailed results are shown in Table~\ref{tab:compare_256_512}. It is evident that the majority of users prefer the stylized results generated by our method.
In other words, although the assessment of stylized results is inherently subjective, our lightweight style transfer agent is designed to generate a diverse array of sequential outputs tailored to meet the varying preferences and requirements of different users. 


\section{Conclusion}

In this paper, we introduce a lightweight Arbitrary Style Transfer method using reinforcement learning. Our approach employs a unified policy to simultaneously learn from content and style images through a coherent encoding and decoding process, thereby more effectively capturing the distinguishing information between content and style. Our novel hierarchical style representation contrastive loss 
differentiates between shallow and deep style representations, enriching the expressiveness of the style transfer. Furthermore, Automatic Multi-task Learning facilitates training across various stages, accelerating the convergence of the model.
Extensive experiments have demonstrated that our method not only generates visually harmonious and aesthetically pleasing artistic images across different resolutions but also produces a diverse range of stylized outcomes. The simplicity and effectiveness of our approach are expected to accelerate the miniaturization of style transfer networks. Although this work has successfully achieved miniaturization and diversification in arbitrary style transfer for images, the challenge remains in applying it to video, which involves temporal processing. Our future goal is to extend our approach to video arbitrary style transfer.

\bibliographystyle{ijcai25}
\bibliography{ijcai25}

\end{document}